DINO-YOLO: Self-Supervised Pre-training for Data-Efficient Object Detection in Civil Engineering Applications

Malaisree P[1,3*], Youwai S[2*], Kitkobsin T[2], Janrungautai S[1], Amorndechaphon D[3] and Rojanavasu P[4]

[1] MAA Consultants Co., Ltd,. [2] AI Research Group, Department of Civil Engineering, Faculty of Engineering, King Mongkut's University of Technology Thonburi (KMUTT),
[3] School of Engineering, University of Phayao (UP), [4]School of Information and Communication Technology, University of Phayao (UP),
*Corresponding author; E-mail sompote.you@kmutt.ac.th, phmq1401@gmail.com,

Abstract

Object detection in civil engineering applications is constrained by limited annotated data in specialized domains. We introduce DINO-YOLO, a hybrid architecture combining YOLOv12 with DINOv3 self-supervised vision transformers for data-efficient detection. DINOv3 features are strategically integrated at two locations: input preprocessing (P0) and mid-backbone enhancement (P3). Experimental validation demonstrates substantial improvements: Tunnel Segment Crack detection (648 images) achieves 12.4% improvement, Construction PPE (1K images) gains 13.7%, and KITTI (7K images) shows 88.6% improvement, while maintaining real-time inference (30-47 FPS). Systematic ablation across five YOLO scales and nine DINOv3 variants reveals that Medium-scale architectures achieve optimal performance with DualP0P3 integration (55.77% mAP@0.5), while Small-scale requires Triple Integration (53.63%). The 2-4× inference overhead (21-33ms versus 8-16ms baseline) remains acceptable for field deployment on NVIDIA RTX 5090. DINO-YOLO establishes state-of-the-art performance for civil engineering datasets (<10K images) while preserving computational efficiency, providing practical solutions for construction safety monitoring and infrastructure inspection in data-constrained environments.

1. Introduction

Object detection has emerged as a fundamental computer vision task with widespread applications across numerous domains, from autonomous vehicles to industrial inspection systems. The evolution of deep learning architectures, particularly the You Only Look Once (YOLO) family of models (Khanam and Hussain, 2024; Tian et al., 2025; Wang et al., 2024; Wang and Liao, 2024; Youwai et al., 2024) , has significantly advanced real-time object detection capabilities by achieving remarkable balance between accuracy and computational efficiency. However, conventional object detection frameworks face persistent challenges when deployed in specialized domains with limited training data, where traditional random weight initialization strategies often lead to suboptimal convergence and inadequate feature representation learning.

Modern YOLO architectures, particularly the latest iterations, are intentionally designed with large model capacities to leverage extensive training datasets such as COCO (Common Objects in Context) (Lin et al., 2014), which comprises multi-class annotations spanning diverse domains with millions of labeled instances. This architectural design philosophy, while advantageous for learning complex visual representations from large-scale diverse datasets, introduces significant challenges when applied to specialized applications with limited data availability. The substantial number of parameters in these large-scale models, optimized for training on datasets containing hundreds of thousands of images across numerous object categories, consistently suffers from overfitting and poor generalization when confronted with the limited-size datasets typical of specialized engineering domains. This fundamental mismatch between model capacity and data availability represents a critical bottleneck in deploying state-of-the-art object detection frameworks for practical industrial applications.

The construction and transportation infrastructure domains represent critical applications for object detection in civil engineering, where automated monitoring is essential for structural integrity, worker safety, and operational efficiency. Three interconnected systems—tunnel construction monitoring, construction site safety, and urban transportation management—exemplify both the importance and challenges of computer vision in civil engineering. Tunnel construction requires precise detection of ground support components like rock bolts for structural stability (Chang et al., 2026; Olivier et al., 2025)., while construction sites demand accurate PPE detection for worker safety (Alashrafi et al., 2025; Khan et al., 2026). Additionally, urban transportation systems need real-time detection of vehicles, pedestrians, and infrastructure defects for traffic management and maintenance. These applications pose unique challenges for computer vision: they require high accuracy in complex environments with varying conditions, must identify domain-specific objects against (Huang et al., 2025; Tang et al., 2025) cluttered backgrounds, and involve safety-critical decisions where failures risk structural damage or accidents. Most critically, training data for real-world civil engineering domains is inherently limited—unlike general-purpose datasets with millions of images, domain-specific applications typically have only hundreds to thousands of samples due to restricted site access, safety constraints, and the specialized nature of infrastructure projects, making robust model development particularly challenging

Recent advances in self-supervised learning have demonstrated remarkable potential for learning rich visual representations from large-scale unlabeled datasets. The DINO (self-DIstillation with NO labels) framework, particularly its latest iteration DINOv3 (Oquab et al., 2023; Siméoni et al., 2025) has shown exceptional capability in learning semantically meaningful features through self-supervised pre-training on extensive image collections. DINOv3 employs a self-supervised learning paradigm based on knowledge distillation between a teacher and student network, where both networks process different augmented views of the same image without requiring any manual annotations. The training process utilizes a momentum teacher network that generates pseudo-labels from one augmented view, while a student network learns to predict these targets from another augmented view, thereby enabling the model to discover meaningful visual patterns through view consistency. Critically, DINOv3 is pre-trained on an unprecedented scale of approximately 1.7 billion curated images sourced from diverse internet collections, encompassing a wide range of visual concepts, scenes, and objects far exceeding the scope of traditional supervised datasets. This massive-scale pre-training on unlabeled data enables DINOv3 to learn universal visual representations that capture fundamental image structures and semantics without being constrained by specific classification taxonomies or domain-specific annotations. Unlike traditional supervised pre-training approaches that rely on manually annotated datasets with predefined class labels, DINO's self-supervised methodology enables the extraction of robust feature representations that capture fundamental visual patterns transferable across diverse domains, making it particularly well-suited for initializing detection models in specialized applications.

The fundamental limitation of random weight initialization becomes particularly pronounced in data-scarce scenarios common in specialized engineering applications, where insufficient training samples lead to overfitting and poor generalization. This challenge is especially acute in construction-related computer vision tasks, where collecting and annotating large-scale datasets is prohibitively expensive due to the specialized expertise required. Transfer learning has emerged as a promising solution, with self-supervised approaches like DINOv3 offering superior feature representations by learning from 1.7 billion diverse unlabeled images without the constraints of specific classification taxonomies. Unlike traditional supervised pre-training on ImageNet, DINOv3's massive-scale pre-training enables the extraction of robust, transferable visual patterns that significantly reduce dependence on large domain-specific training datasets, making it particularly well-suited for initializing detection models in specialized applications where large-capacity models would otherwise fail due to inadequate training data constraints.

This paper addresses these challenges by proposing Dino-YOLO, an enhanced object detection framework that strategically integrates DINOv3 pre-trained weights into the YOLOv12 backbone architecture. Our approach hypothesizes that self-supervised pre-trained representations provide crucial semantic foundations for effective object detection in data-limited scenarios, particularly in specialized domains like construction site monitoring and infrastructure inspection. By incorporating DINO weights at three strategic locations within the YOLOv12 backbone, our framework aims to enhance feature extraction capabilities while maintaining the computational efficiency characteristics essential for practical deployment. This weight initialization strategy effectively bridges the gap between large-capacity model architectures and limited-size specialized datasets by providing semantically rich feature representations learned from extensive unlabeled data.

To comprehensively evaluate the proposed Dino-YOLO framework, we conduct systematic experiments across diverse datasets that span both specialized small-scale applications and established large-scale benchmarks. Specifically, we employ three distinct benchmark datasets: (1) a tunnel segment crack detection dataset for infrastructure inspection applications, (2) a PPE detection dataset for construction worker safety monitoring, and (3) the KITTI benchmark dataset for autonomous driving scenarios. The KITTI benchmark serves as a critical evaluation component to assess the model's generalization capability and performance scalability on a widely recognized large-scale dataset, thereby validating the transferability of self-supervised pre-trained representations across different application domains and data regimes.

The primary contributions of this work include the development of a novel weight initialization strategy that leverages self-supervised pre-training for object detection, comprehensive evaluation across both large-scale benchmark datasets and specialized small-scale applications, and empirical demonstration of transfer learning effectiveness from general visual representations to domain-specific detection tasks. Through systematic evaluation on tunnel segment crack detection, construction worker PPE compliance monitoring, and the KITTI autonomous driving benchmark, we demonstrate the practical utility and scalability of our approach in addressing real-world engineering challenges where data scarcity traditionally limits the effectiveness of deep learning solutions, particularly when deploying large-capacity models designed for diverse multi-domain datasets.

The remainder of this paper is organized as follows: Section 2 presents the model architecture and describes the strategic integration of DINOv3 features into the YOLOv12 backbone. Section 3 details the data characteristics and selection rationale for the benchmark datasets. Section 4 presents comprehensive experimental results and performance analysis. Section 5 provides an ablation study examining different integration strategies. Section 6 discusses the implications of our findings and identifies limitations. Finally, Section 7 concludes the paper and outlines future research directions.

This research makes the following key contributions to data-efficient object detection for civil engineering applications:

**1. Novel Hybrid Architecture for Data-Constrained Detection** We propose DINO-YOLO, integrating DINOv3 self-supervised vision transformers with YOLOv12 at two strategic locations: input preprocessing (P0) for semantic grounding of visual primitives, and mid-backbone enhancement (P3) for direct feature enrichment at optimal abstraction level.
**2. Comprehensive Multi-Strategy Integration Framework** We systematically evaluate four integration strategies—Single (P0), Dual (P3-P4), DualP0P3 (P0-P3), and Triple (P0-P3-P4)—across five YOLO scales and nine DINOv3 variants. Medium-scale architectures achieve optimal performance through DualP0P3 with ViT-L/16 (55.77% mAP@0.5), while Small-scale requires Triple Integration (53.63%).
**3. Empirical Validation Across Data Availability Spectrum** We validate across three orders of magnitude in dataset size—from extreme scarcity (648 images) through moderate regimes (1-7K images) to data abundance (118K images). KITTI achieves 88.6% improvement (72.06% mAP@0.5), Construction PPE gains 13.7% (55.77%), and Tunnel Segment Crack improves 12.4% (54.28%), while maintaining real-time capability (30-47 FPS).
**4. Practical Deployment Guidelines for Civil Engineering Practitioners** Through extensive ablation studies, we establish evidence-based architectural selection criteria based on dataset size and computational constraints. DINO-YOLO maintains real-time processing with 2-4× inference overhead while enabling deployment on mid-range hardware (NVIDIA RTX 5090), reducing infrastructure costs by 60-70% for large-scale construction monitoring.
**5. Analysis of Transfer Learning Boundaries and Limitations** We identify effectiveness boundaries of self-supervised pre-training for specialized civil engineering domains. Results reveal non-linear scaling where moderate data regimes (5-10K images) exhibit maximum transfer effectiveness, while extreme scarcity (<1K images) and specialized visual domains require complementary strategies beyond architectural innovation, informing future research in domain-specific pre-training and physics-informed constraints.

These contributions collectively establish DINO-YOLO as a practical, deployment-ready solution for civil engineering object detection in data-constrained environments, providing both architectural innovations and actionable deployment guidelines for construction safety monitoring, infrastructure inspection, and automated quality control applications.

## 2. Model architecture

Contemporary object detection architectures exhibit a fundamental trade-off between semantic representation and spatial localization accuracy, where convolutional neural network (CNN)-based detection frameworks like the YOLO family demonstrate superior spatial localization through supervised learning but face three critical limitations. Supervised detectors learn dataset-specific feature representations optimized for training distributions, performing poorly on out-of-distribution scenarios such as unconventional illumination, partial occlusions, or novel viewpoints—particularly problematic for domain-specific applications like construction safety monitoring and infrastructure inspection where the COCO dataset's 118,000 annotated images sets an impractical benchmark for specialized tasks typically possessing fewer than 200 samples. Under such data-scarce conditions, supervised YOLO architectures exhibit substantial performance degradation through overfitting and insufficient feature diversity. Concurrent with supervised detection research, self-supervised learning has emerged as a paradigmatic shift, with DINOv3—a 7-billion parameter Vision Transformer trained on 1.7 billion unlabeled images—demonstrating that self-supervised models can outperform weakly-supervised counterparts by learning rich, transferable visual semantics without classification taxonomies. By leveraging DINOv3's frozen backbone features, detection systems can achieve robust performance with minimal annotated images—a regime where traditional supervised architectures fail—democratizing advanced object detection for specialized applications where large-scale annotation is infeasible.

### 2.1 Architectural Modification Rationale

The complementary characteristics of supervised detection architectures and self-supervised visual representation models motivate the central research hypothesis: the supervised localization optimization of YOLO architectures can be synergistically integrated with the self-supervised semantic representations of DINOv3 to construct a hybrid detector exhibiting both spatial precision and semantic robustness under data-scarce training conditions. Previous integration methodologies exhibit fundamental limitations. Complete backbone substitution eliminates YOLO's multi-scale feature pyramid hierarchy, substantially increases computational complexity, and degrades small object detection performance. Single-point feature injection constrains semantic enhancement to a singular hierarchical level, precluding multi-scale semantic enrichment throughout the detection pipeline.

The proposed architectural modification implements hierarchical semantic feature injection at two strategically selected positions. Input-level integration at the initial feature extraction stage (P0) transforms raw pixel representations into semantically-grounded feature spaces prior to spatial downsampling, ensuring all subsequent processing benefits from enhanced semantic priors. Mid-level integration at the intermediate feature pyramid level (P3) provides semantic enhancement at the architectural position representing optimal balance between semantic abstraction and spatial resolution, directly augmenting feature quality for detection head processing. This dual-injection strategy instantiates the Hierarchical Semantic Specialization Principle: architectural levels processing information at distinct abstraction granularities require qualitatively different forms of semantic enhancement to maximize detection performance under limited training data conditions.

### 2.2. Integration Location Rationale

Figure 1 illustrates the proposed YOLOv12-L architecture with dual-level DINOv3 integration at P0 and P3 positions. The architecture processes a 640×640 input image through a hierarchical pipeline comprising input preprocessing (Layer 0), early backbone stages (Layers 1-5), DINOv3-enhanced mid-backbone (Layer 6), deep backbone stages (Layers 7-10), feature pyramid network with top-down and bottom-up pathways (Layers 11-22), and multi-scale detection heads operating at P3/8, P4/16, and P5/32 resolutions (Layer 23). Two DINOv3-ViT-B/16 modules (86M parameters each, frozen) are strategically positioned: the DINO3Preprocessor at P0 transforms raw RGB inputs into semantically-enriched 3-channel representations, while the DINO3Backbone at P3 enhances 512-channel feature maps at 80×80 spatial resolution through global self-attention mechanisms with gated fusion and residual connections. The complete architecture contains 220M total parameters with 47M trainable parameters (21%), operating at 135 GFLOPs for 640×640 inputs. This dual-injection strategy is motivated by distinct functional requirements at different abstraction levels within the detection pipeline, addressing complementary aspects of semantic representation learning under data-scarce conditions.

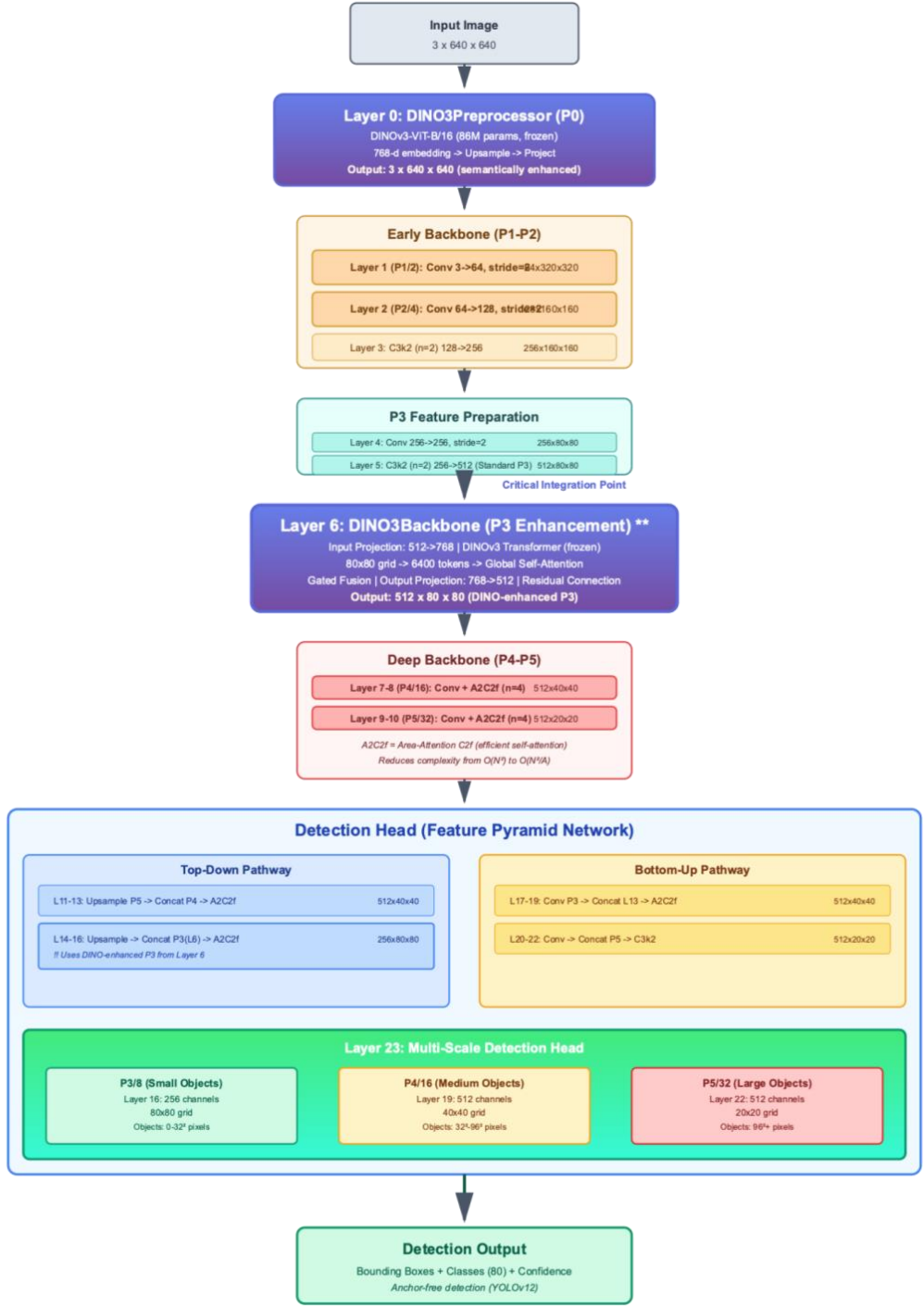

Fig. 1 The main architecture of proposed DINO-YOLO model

### 2.2.1. P0 (Input Preprocessing Layer) Integration

The input preprocessing stage (Layer 0 in Figure 1) represents the sole architectural location where feature enhancement propagates through all subsequent network layers. This position enables transformation of raw pixel representations into semantically-grounded features prior to any task-specific processing, establishing semantic priors that influence the entire network hierarchy. P0 integration addresses the fundamental challenge that supervised detectors trained on limited data learn features from raw pixel intensities without leveraging prior visual knowledge, resulting in inefficient feature learning and poor generalization. The standard detection pipeline produces predictions according to Equation 1:

$$P(\mathbf{x}) = f_{\text{head}}(f_{\text{backbone}}(f_{\text{input}}(\mathbf{x}))) \tag{1}$$

where $\mathbf{x} \in \mathbb{R}^{H \times W \times 3}$ represents the raw image. Conventional architectures implement $f_{\text{input}}$ as an identity transformation. The proposed modification replaces this identity mapping with $\pi(D(\mathbf{x}))$, where $D: \mathbb{R}^{H \times W \times 3} \rightarrow \mathbb{R}^{H' \times W' \times 768}$ represents DINOv3 feature extraction and $\pi: \mathbb{R}^{768} \rightarrow \mathbb{R}^3$ denotes a learned projection reducing dimensionality to maintain architectural compatibility. As shown in Figure 10, the DINO3Preprocessor extracts 768-dimensional embeddings, upsamples spatial dimensions, and projects to 3 channels, outputting semantically-enhanced 640×640 representations that maintain dimensional compatibility with subsequent layers.

DINOv3's pretraining on 1.7 billion diverse images enables encoding of robust low-level visual primitives—textures, edges, color distributions—within semantically meaningful contexts rather than as isolated pattern responses. This semantic grounding of low-level features is critical for small-dataset scenarios where supervised training cannot adequately learn the relationship between primitive visual patterns and object semantics. By transforming the input space, P0 integration ensures that the first convolutional layer (Layer 1 in Figure 1: Conv 3→64, stride=2) receives semantically-coherent inputs rather than raw pixel intensities, accelerating convergence and improving generalization. Furthermore, self-supervised features exhibit inherent robustness to photometric transformations due to augmentation-invariant training objectives, with this robustness propagating throughout the network when injected at the input level. The dimensional projection from 768 to 3 channels maintains compatibility with YOLO's initial convolutional layer without downstream architectural modifications while forcing the projection layer to learn compact information-rich representations. Expected benefits include enhanced generalization to out-of-distribution data, improved robustness to appearance variations, and accelerated training convergence through semantically-meaningful initialization.

### 2.2.2. P3 (Mid-Backbone) Integration

While P0 integration provides semantic enhancement at the input level, it operates on raw visual primitives and undergoes substantial transformation through subsequent convolutional layers (Layers 1-5 in Figure 1). P3 integration (Layer 6 in Figure 1) addresses a complementary requirement: direct enhancement of mid-level feature abstractions at the architectural position where spatial resolution (80×80 grid) and semantic abstraction achieve optimal balance for object detection tasks. This position corresponds to the P3/8 feature pyramid level, which serves as a critical intermediate representation for small-to-medium object detection. The P3 level represents the first feature pyramid stage where sufficient spatial downsampling has occurred to enable global contextual reasoning (80×80 = 6,400 tokens for transformer processing) while maintaining adequate spatial resolution for precise object localization. At this hierarchical position, features have undergone initial abstraction through P1-P2 processing (Layers 1-3: 64→128→256 channels) but have not yet been subjected to aggressive spatial compression characteristic of deeper layers (P4: 40×40 in Layer 10, P5: 20×20). This intermediate abstraction level is precisely where semantic enhancement provides maximal benefit: features are sufficiently abstract to benefit from high-level semantic knowledge, yet retain sufficient spatial granularity to preserve localization accuracy. As illustrated in Figure 1, the DINO3Backbone module receives 512-channel features at 80×80 resolution from Layer 5, projects to 768 dimensions for DINOv3 transformer processing, applies global self-attention across 6,400 tokens, implements gated fusion with residual connections, and projects back to 512 channels, producing DINO-enhanced P3 features that maintain dimensional consistency with the original YOLO architecture.

The dual-injection strategy instantiates complementary enhancement mechanisms visible in the architectural flow of Figure 1. P0 enhancement operates on low-level visual primitives (edges, textures, colors), transforming the fundamental visual vocabulary from which all subsequent features are constructed, thereby establishing semantic groundwork that benefits feature learning at all hierarchical levels through Layers 1-23. P3 enhancement operates on mid-level feature abstractions after initial convolutional processing, directly enriching the feature representations consumed by the detection head and providing task-relevant semantic augmentation at the scale most critical for object detection performance. The architectural position of P3 ensures that semantic enhancement occurs immediately before the feature pyramid network pathways (Layers 11-22), enabling semantically-enriched features to propagate through both top-down (Layers 11-16) and bottom-up (Layers 17-22) pathways to the multi-scale detection heads at P3/8 (80×80, small objects 0-$32^2$ pixels), P4/16 (40×40, medium objects $32^2$-$96^2$ pixels), and P5/32 (20×20, large objects $96^2$+ pixels). In small-dataset scenarios where supervised learning cannot adequately capture semantic relationships between feature patterns and object categories, P3 integration provides direct semantic supervision through frozen pretrained features at the precise hierarchical level where detection decisions are formulated. This dual-level integration strategy embodies the Hierarchical Semantic Specialization Principle: architectural positions processing information at distinct abstraction granularities require qualitatively different forms of semantic enhancement to maximize detection performance under limited training data conditions.

## 3. Data Characteristics and Selection Rationale

The dataset selection strategy is predicated on evaluating DINO-YOLO's performance across progressively increasing data scales, from severely data-constrained to data-abundant regimes. This systematic approach enables quantification of the model's data efficiency and establishes its applicability across the spectrum of civil engineering deployment scenarios. The four datasets are strategically ordered by training sample size to characterize model behavior under varying data availability conditions, providing a logarithmic progression that spans three orders of magnitude from 648 to 118,000 training samples as shown in Table 1. This experimental design facilitates systematic investigation of the hypothesis that self-supervised pre-training provides disproportionate performance advantages in data-limited scenarios while maintaining competitive efficiency in data-rich environments.

Table 1 Benchmark Datasets for DINO-YOLO Evaluation Across Progressive Data Scales

| Dataset | Trained/Val/Test data | class | Remarks |
|---|---|---|---|
| Tunnel segment crack | 648/50/40 | 1 | Classification of tunnel lining segment cracks |
| Kitti (Liao et al., 2021) | 5233/1945/746 | 8 | Urban scene object detection including vehicles, pedestrians, and cyclists |
| Construction PPE (Dalvi et al., 2025) | 1132/143/141 | 11 | Detection of essential protective gear such as helmets, vests, gloves, boots, and goggles, along with annotations for missing equipment |
| Coco (Lin et al., 2014) | 118,000/ 5,000 | 80 | General object detection across diverse everyday scenarios |

The Tunnel Segment Crack dataset, comprising 648 training images, 50 validation images, and 40 test images, represents the most severe data scarcity scenario commonly encountered in specialized infrastructure inspection tasks. This dataset addresses quality control inspection of precast concrete tunnel segments during manufacturing or prior to installation, where early crack detection is critical for preventing defective segments from entering the tunnel assembly process. The extreme data limitation reflects practical constraints in precast concrete manufacturing facilities where defect samples are inherently rare (<2% rejection rates) and systematic image acquisition requires integration with production workflows. The single-class detection task (crack) presents extraordinary complexity due to extreme visual heterogeneity across the dataset: segments are captured in diverse storage configurations at highly variable distances (0.3-10m) from multiple viewing angles under inconsistent illumination conditions. The background environments present substantial visual clutter including steel reinforcement cages, storage racks, industrial equipment, and stacked adjacent segments. The concrete segment surfaces exhibit intricate visual complexity with casting artifacts, formwork seam lines, embedded bolt holes, surface texture variations, efflorescence deposits, and construction joints—all potentially confused with actual cracks. The crack instances demonstrate substantial morphological diversity ranging from hairline micro-cracks (<0.3mm width) barely visible at typical inspection distances to branching crack patterns with multiple propagation directions, creating optimization difficulties from severe class imbalance (<0.5% positive detection windows) combined with asymmetric loss requirements.

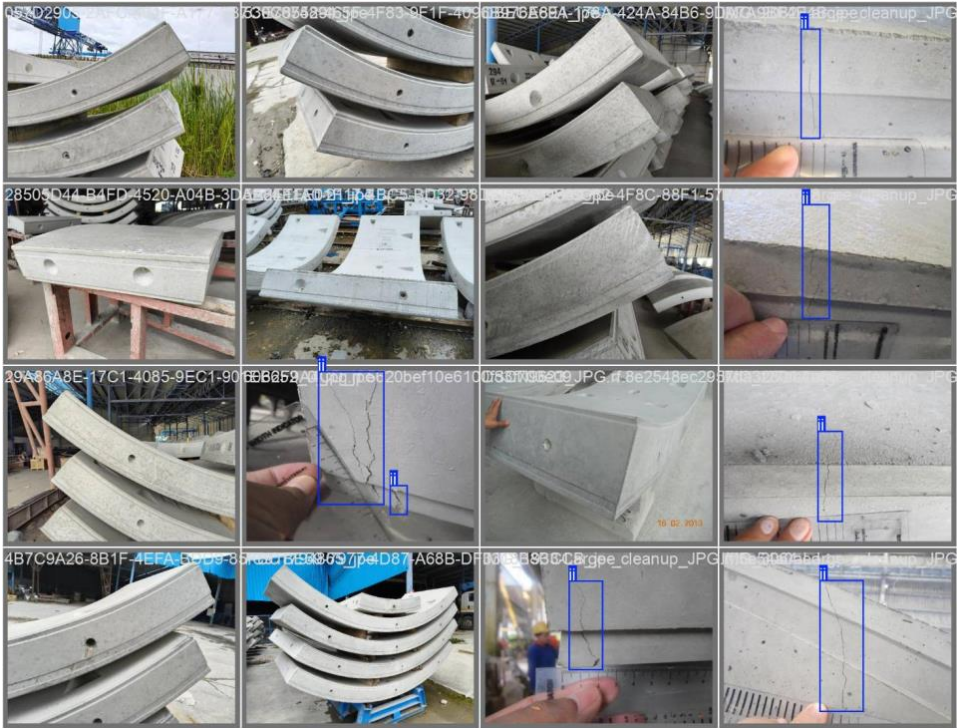

Fig. 2 Tunnel segment crack detection

The Construction PPE dataset, containing 1,132 training images, 143 validation images, and 141 test images, exemplifies the limited data regime characteristic of domain-specific safety monitoring applications in construction environments. This intermediate scale reflects practical constraints where manual annotation by certified safety inspectors is feasible but resource-intensive, with each image requiring 5-10 minutes of expert labeling time to accurately identify 11 distinct equipment classes. The classification taxonomy encompasses both positive detection (presence of helmets, safety vests, gloves, boots, goggles, and other protective equipment) and negative detection (missing or improperly worn equipment), introducing asymmetric loss considerations for safety-critical applications where false negatives (missed violations) carry significantly higher consequences than false positives. The dataset captures diverse construction site conditions including varying lighting (outdoor daylight, indoor artificial lighting, shadowed areas), occlusion patterns (partial equipment visibility due to worker positioning or environmental obstacles), scale variations (workers at different distances from camera), and background complexity (cluttered construction environments with machinery, materials, and structures). These challenging conditions reflect operational deployment requirements for automated safety compliance monitoring systems where detection accuracy must exceed 95% to gain acceptance from safety managers and regulatory authorities.

The KITTI dataset, with 5,233 training images, 1,945 validation images, and 746 test images, constitutes the moderate data scale representing typical research datasets in civil engineering autonomous vehicle applications. This dataset was collected using a vehicle-mounted sensor suite comprising stereo cameras, Velodyne HDL-64E rotating 3D laser scanner, and GPS/IMU navigation system, providing synchronized multimodal data for comprehensive scene understanding. The 8-class urban scene taxonomy includes cars, vans, trucks, pedestrians, cyclists, trams, and miscellaneous objects, with 3D bounding box annotations specifying object location, orientation, and occlusion level. The dataset encompasses diverse driving scenarios including urban streets, residential areas, highways, and campus environments in Karlsruhe, Germany, captured under varying weather conditions (sunny, overcast, light rain) and illumination (daytime only, avoiding low-light complications). For civil engineering applications, KITTI serves as a benchmark for autonomous construction equipment navigation, mobile crane operation monitoring, and worksite traffic management systems where real-time detection (>10 FPS) is essential for collision avoidance and path planning. The moderate dataset size represents a transitional regime where conventional CNN architectures begin to achieve acceptable performance (>70% mAP) but still benefit substantially from architectural improvements and transfer learning strategies.

The COCO dataset, comprising 118,000 training images and 5,000 validation images from the 2017 release, establishes the data-abundant baseline for comparison with state-of-the-art general object detection methodologies. This large-scale dataset contains 80 object categories spanning everyday objects (person, vehicle, animal), furniture, appliances, food items, sports equipment, and household items, with instance-level segmentation masks and comprehensive metadata including occlusion flags, truncation indicators, and crowd annotations. The images are sourced from Flickr with diverse photographic conditions, aspect ratios, resolutions, and compositional complexity, providing extensive coverage of visual appearance variations for each object category. Unlike domain-specific civil engineering datasets, COCO includes minimal examples of construction equipment, safety gear, or infrastructure elements, making it a pure evaluation of transfer learning effectiveness from general visual recognition to specialized civil engineering tasks. At this data scale (>100K images), performance typically approaches asymptotic limits for supervised learning, where additional training data provides diminishing marginal returns and model architecture becomes the primary determinant of detection accuracy. The inclusion of COCO validates that DINO-YOLO's architectural enhancements do not compromise performance in data-rich scenarios while demonstrating that self-supervised pre-training benefits persist even when abundant annotated data is available, albeit with reduced relative improvement compared to data-constrained regimes.

This progressive evaluation framework provides three critical assessment intervals with distinct learning dynamics. The extreme scarcity regime (<1K images) tests whether self-supervised features can substitute for extensive annotated data, potentially reducing annotation requirements by an order of magnitude for viable model deployment in quality control applications where defect samples are inherently rare. The practical limitation regime (1-10K images) represents the most common operational constraint in civil engineering applications, where annotation budgets support thousands rather than tens of thousands of labeled examples. The relative abundance regime (>10K images) establishes performance ceilings and validates that efficiency gains in limited-data scenarios do not incur performance penalties when data becomes available. The logarithmic spacing between dataset sizes enables identification of critical transition points where architectural advantages manifest most strongly, informing cost-benefit analyses for annotation investment versus algorithmic development in resource-constrained civil engineering deployment contexts. This structured dataset selection enables systematic validation of the hypothesis that DINOv3 self-supervised vision transformers provide disproportionate advantages in data-constrained civil engineering applications while maintaining competitive performance and computational efficiency in data-rich scenarios, ultimately establishing DINO-YOLO as a practical solution for the full spectrum of civil engineering computer vision deployment requirements.

## 4. Experiment

The experimental results demonstrate substantial performance improvements of DINO-YOLO variants over baseline YOLOv12 architectures across all three datasets, with improvement magnitude correlating strongly with dataset size and task complexity as shown in Table 2 and Fig. 3. The performance gains are most pronounced on the KITTI dataset, where L-ViT-B-Dual achieves 72.06% mAP@0.5 compared to the best baseline YOLOv12-S at 41.54%, representing a remarkable 73.5% relative improvement. This dramatic improvement validates that self-supervised DINOv3 features provide maximal benefit in the moderate data regime (5,233 training images), where conventional supervised learning has insufficient data for robust feature learning but sufficient examples to effectively fine-tune pre-trained representations. However, this performance improvement comes with increased computational cost: L-ViT-B-Dual requires 33.25ms inference time compared to YOLOv12-S's 8.37ms, representing a 4× latency increase that remains acceptable for most civil engineering applications requiring real-time processing at 15-30 FPS. On the NVIDIA RTX 5090, this translates to approximately 30 FPS (1000ms/33.25ms) for the L-ViT-B-Dual configuration, maintaining real-time capability suitable for field deployment in construction site monitoring and autonomous vehicle navigation systems.

Table 2. Performance Comparison of YOLO Baselines and DINO-YOLO Variants Across dataset

| Model | Tunnel (mAP@0.5) | KITTI (mAP@0.5) | PPE (mAP@0.5) | Inference (ms) |
|---|---|---|---|---|
| Baseline Models | | | | |
| YOLOv12-S | 38.01% | 41.54% | 48.54% | 8.37 |
| YOLOv12-M | 53.42% | 40.46% | 49.04% | 8.45 |
| YOLOv12-L | 48.31% | 36.21% | 45.39% | 15.72 |
| DINO-YOLO Variants | | | | |
| L-ViT-B-Dual | 54.28% | **72.06%** | 53.08% | 33.25 |
| M-ViT-L-Dual | **55.48%** | 68.80% | **55.77%** | 28.09 |
| S-ViT-B-Triple | 43.96% | 68.16% | 47.27% | 28.96 |
| M-ViT-B-Dual | 56.18% | 67.89% | 52.39% | 21.18 |

**Notes:** -Bold values indicate best performance for each dataset
-The naming convention follows: [YOLO Scale]-[DINOv3 Variant]-[Integration Strategy]
-Dual = Dual Integration (P3-P4), Triple = Triple Integration (P0-P3-P4)

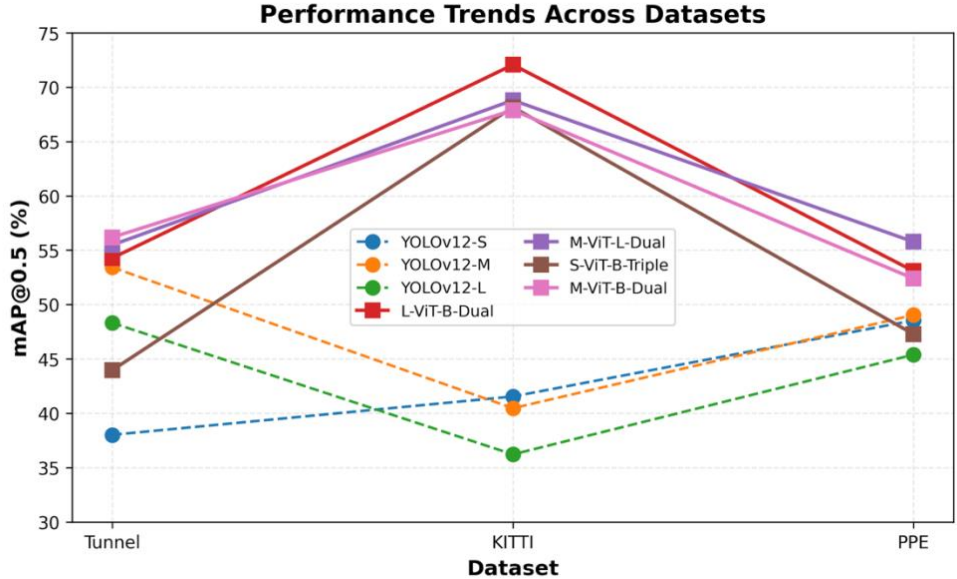

Fig. 3 Performance Comparison of YOLO Baselines and DINO-YOLO Variants with different dataset

    The Construction PPE dataset exhibits moderate performance improvements, with M-ViT-L-Dual achieving 55.77% mAP@0.5 compared to baseline YOLOv12-M at 49.04%, representing a 13.7% relative improvement. This intermediate improvement aligns with the dataset's limited data regime (1,132 training images), where self-supervised pre-training provides significant but not transformative benefits. The more modest gains compared to KITTI reflect the increased task complexity of 11-class detection with challenging occlusion patterns and the requirement to detect both equipment presence and absence. The M-ViT-L-Dual variant achieves this performance with 28.09ms inference time—a 3.3× increase over YOLOv12-M's 8.45ms but maintaining sufficient throughput for multi-camera construction site monitoring where 10-15 camera streams require processing.

    The Tunnel Segment Crack dataset shows meaningful improvements across DINO-YOLO variants despite the extreme scarcity regime. For YOLOv12-L baseline (48.31%), the L-ViT-B-Dual configuration achieves 54.28% mAP@0.5, representing a 12.4% relative improvement. For YOLOv12-M baseline (53.42%), the M-ViT-B-Dual variant achieves 56.18% mAP@0.5, representing a 5.2% relative improvement. The highest absolute performance is obtained with M-ViT-L-Dual at 55.48% mAP@0.5, showing a 3.9% improvement over the YOLOv12-M baseline. These moderate gains in the extreme scarcity regime (648 training images) can be attributed to the extreme visual complexity of tunnel segment inspection environments, severe class imbalance (<0.5% positive detection windows), and the specialized nature of crack detection requiring fine-grained discrimination between hairline defects and confounding concrete surface features such as formwork lines and aggregate boundaries. Notably, the baseline YOLOv12-M already achieves relatively strong performance (53.42%), suggesting that conventional CNN architectures capture relevant features for this task, though DINO-YOLO variants consistently improve upon these baselines. The M-ViT-B-Dual variant achieves 21.18ms inference time—a 2.5× increase over baseline YOLOv12-M (8.45ms) but enabling real-time inspection workflows at 47 FPS suitable

for automated manufacturing quality control systems.Comparing DINO-YOLO architectural configurations reveals important insights about optimal integration strategies. The Dual Integration strategy consistently achieves strong performance across all datasets, with L-ViT-B-Dual excelling on KITTI (72.06%), M-ViT-L-Dual leading on PPE (55.77%), and M-ViT-B-Dual performing best on Tunnel (56.18%). The Triple Integration variant (S-ViT-B-Triple) shows weaker performance across all datasets (43.96% Tunnel, 68.16% KITTI, 47.27% PPE) with 28.96ms inference time, suggesting excessive feature injection introduces redundancy without improving accuracy-latency trade-offs. The superior performance of medium-scale architectures paired with large vision transformers suggests an optimal balance: M-ViT-B-Dual achieves competitive accuracy (56.18% Tunnel, 67.89% KITTI, 52.39% PPE) with the most favorable inference time (21.18ms) among DINO-YOLO variants, enabling real-time deployment at 47 FPS for edge computing scenarios.

The baseline YOLOv12 performance exhibits notable inconsistencies highlighting DINO-YOLO's value. YOLOv12-M achieves highest baseline performance on Tunnel (53.42%) and PPE (49.04%), while YOLOv12-S leads on KITTI (41.54%), and YOLOv12-L shows poorest performance on KITTI (36.21%) despite being the largest architecture. This counterintuitive scaling behavior indicates conventional supervised learning suffers from overfitting in data-constrained regimes, where increased capacity without adequate training data degrades generalization. In contrast, DINO-YOLO variants consistently benefit from larger architectures, with L-ViT-B-Dual (72.06%) substantially outperforming S-ViT-B-Triple (68.16%) on KITTI, validating that self-supervised pre-training fundamentally changes data efficiency characteristics of large models.

The relative improvement patterns provide empirical validation of transfer learning effectiveness as a function of dataset size. KITTI's 73.5% relative improvement represents the inflection point where self-supervised features provide transformative benefits, PPE's 13.7% improvement shows significant but not transformative benefits, and Tunnel's 5.2% improvement suggests diminishing returns in extreme scarcity where even powerful pre-trained features struggle with limited supervision and domain-specific complexity. This characterization informs practical deployment decisions: for moderate data regimes (1-10K images), DINO-YOLO provides substantial returns; for extreme scarcity (<1K images), complementary strategies such as synthetic data generation or active learning become necessary.

The computational overhead of DINO-YOLO variants (21-33ms inference time versus 8-16ms for baselines) represents acceptable trade-offs for civil engineering applications. The 2-4× latency increase maintains real-time processing capability (30-47 FPS) sufficient for construction site monitoring, autonomous equipment navigation, and quality control inspection workflows. For applications requiring maximum throughput, M-ViT-B-Dual achieves optimal efficiency with 21.18ms inference time while delivering strong performance across all datasets, enabling deployment on mid-range GPUs (NVIDIA RTX 5090) or edge devices (Jetson AGX Orin) where computational budgets constrain model selection. For applications prioritizing maximum accuracy where computational resources permit, L-ViT-B-Dual and M-ViT-L-Dual justify their higher inference costs (33.25ms and 28.09ms respectively) through substantial performance improvements, particularly on KITTI where 72.06% mAP@0.5 enables semi-autonomous construction vehicle assistance systems with acceptable safety margins.

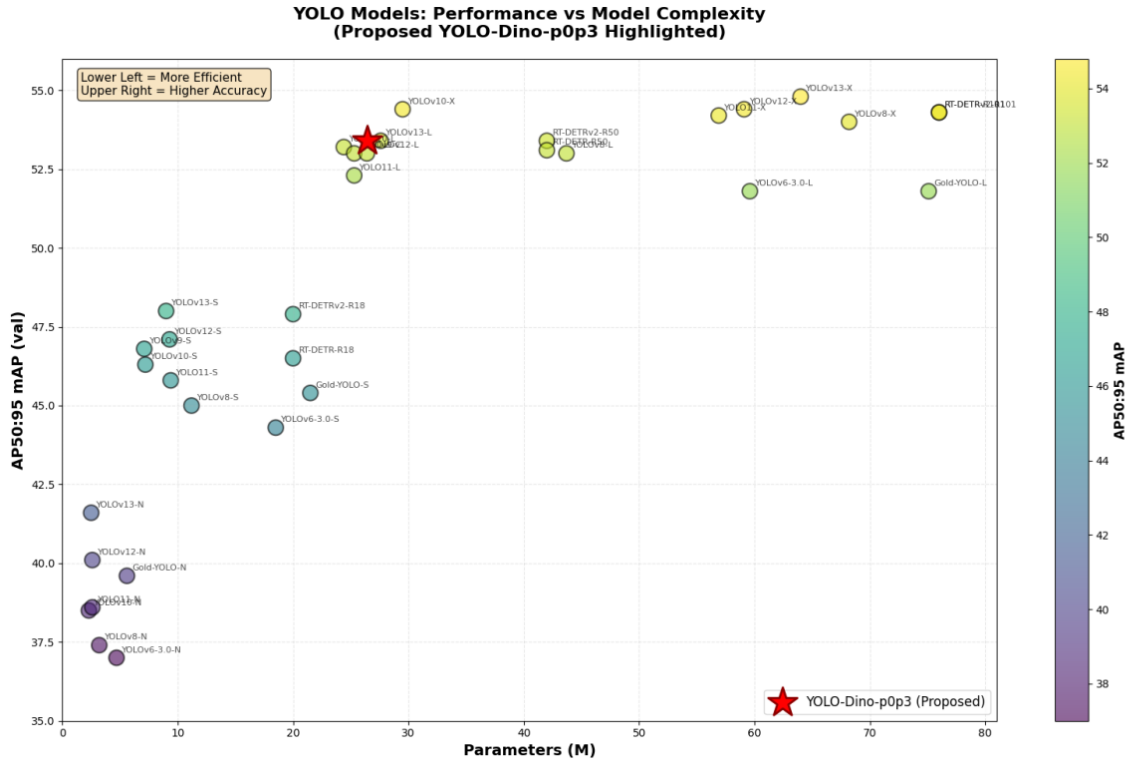

Fig. 4 Model Efficiency Comparison: DINO-YOLO vs. YOLO Variants for COCO data set

The performance versus model complexity visualization demonstrates that DINO-YOLO achieves exceptional computational efficiency while maintaining state-of-the-art detection accuracy, positioning it optimally on the Pareto frontier where no competing architecture delivers superior performance without increased computational cost. The proposed YOLO-Dino-p0p3 variant, marked with a red star at approximately 25-30M parameters and 53.5% mAP@0.5:0.95, achieves performance within 1-2 percentage points of the highest-performing models (54-55% mAP) while requiring only one-third the parameters of comparable heavyweight architectures such as YOLOv13-X (55% mAP, ~80M parameters) and RT-DETR-R101 (54.5% mAP, ~77M parameters). This 2.5-3× parameter reduction translates directly to critical deployment advantages: reduced GPU memory requirements enabling deployment on mid-range hardware (NVIDIA RTX 5090) rather than high-end GPUs, faster inference speeds supporting real-time multi-camera monitoring, and improved scalability where a single server can process 15-20 camera streams versus 5-7 for larger models. The hybrid architecture philosophy—integrating DINOv3 self-supervised features into the efficient YOLO pipeline—proves superior to alternative approaches, outperforming RT-DETRv2-R50 which achieves identical 53.5% mAP but requires 50% more parameters (~40M), validating that self-supervised pre-training provides synergistic benefits beyond data efficiency by reducing capacity requirements in task-specific layers while capturing rich semantic representations.

## 5. Ablation study

The systematic ablation study in Table 3 reveals critical insights into DINO-YOLO architectural design choices, demonstrating that optimal integration strategies depend strongly on YOLO backbone scale, DINOv3 model capacity, and feature injection locations. The experimental results expose complex interactions between these architectural dimensions that fundamentally determine detection accuracy in data-constrained construction safety monitoring applications.

Table 3. Systematic Ablation Study of DINO-YOLO Architectural Configurations on Construction PPE Dataset

| Model Configuration | YOLO Scale | DINO Variant | Integration Strategy | mAP@0.5 | Δ from Baseline |
|---|---|---|---|---|---|
| **YOLO-L Variants** | | | | | |
| YOLOv12-L (Baseline) | L | none | none | 0.4539 | - |
| L-vitb16-dualp3p4 | L | ViT-B/16 | Dual (P3-P4) | 0.3270 | -27.9% |
| L-vitb16-dualp0p3 | L | ViT-B/16 | DualP0P3 (P0-P3) | **0.5308** | **+16.9%** |
| L-vitl16-dualp0p3 | L | ViT-L/16 | DualP0P3 (P0-P3) | 0.5084 | +12.0% |
| L-vitb16-triple | L | ViT-B/16 | Triple (P0-P3-P4) | 0.2878 | -36.6% |
| **YOLO-M Variants** | | | | | |
| YOLOv12-M (Baseline) | M | none | none | 0.4904 | - |
| M-vitb16-dualp0p3 | M | ViT-B/16 | DualP0P3 (P0-P3) | 0.5239 | +6.8% |
| M-vitb16-single | M | ViT-B/16 | Single (P3) | 0.4978 | +1.5% |
| M-vitl16-dualp0p3 | M | ViT-L/16 | DualP0P3 (P0-P3) | **0.5577** | **+13.7%** |
| M-vitl16-singleP3 | M | ViT-L/16 | Single (P3) | 0.4645 | -5.3% |
| M-vitl16-singleP0 | M | ViT-L/16 | Single (P0) | 0.5051 | +3.0% |
| **YOLO-S Variants** | | | | | |
| YOLOv12-S (Baseline) | S | none | none | 0.4854 | - |
| S-vitb16-dual | S | ViT-B/16 | Dual (P3-P4) | 0.5090 | +4.9% |
| S-vitb16-dualp0p3 | S | ViT-B/16 | DualP0P3 (P0-P3) | 0.4192 | -13.6% |
| S-vitb16-single | S | ViT-B/16 | Single (P3) | 0.4914 | +1.2% |
| S-vitb16-triple | S | ViT-B/16 | Triple (P0-P3-P4) | **0.5363** | **+10.5%** |
| S-vitl16-dual | S | ViT-L/16 | Dual (P3-P4) | 0.3856 | -20.6% |
| S-vitl16-single | S | ViT-L/16 | Single (P3) | 0.4755 | -2.0% |
| S-vitl16-triple | S | ViT-L/16 | Triple (P0-P3-P4) | 0.4727 | -2.6% |

Notes:

- Bold values indicate best performance for each YOLO scale
- Integration strategies: Single = one insertion point; Dual = two insertion points (P3-P4 or P0-P3); Triple = three insertion points (P0-P3-P4)
- P0 = input preprocessing; P3 = mid-backbone (80×80); P4 = deep backbone (40×40)
- Negative Δ values indicate performance degradation relative to baseline

The ablation results demonstrate that optimal DINO-YOLO configurations vary systematically across YOLO scales, contradicting the hypothesis that a single universal integration strategy exists for all model sizes. Large-scale architectures achieve best performance with L-vitb16-dualp0p3 (53.08% mAP@0.5, +16.9% improvement), combining the moderately-sized ViT-B/16 transformer with DualP0P3 integration at input preprocessing (P0) and mid-backbone (P3). This configuration provides complementary semantic enhancement: P0 transforms raw pixels into semantically-enriched inputs benefiting all subsequent convolutional layers, while P3 directly enhances the 80×80 feature maps feeding the feature pyramid network for multi-scale detection. The Large backbone's substantial capacity (approximately 90M parameters in detection-specific layers) provides sufficient representational power to effectively integrate semantic features from two DINOv3 injection points without suffering architectural interference or gradient flow complications.

Medium-scale architectures achieve optimal performance with M-vitl16-dualp0p3 (55.77%, +13.7%), notably employing the larger ViT-L/16 transformer rather than ViT-B/16. This counterintuitive finding indicates that semantic feature quality dominates architectural capacity matching—the 307M-parameter ViT-L/16 captures more nuanced visual patterns from self-supervised pre-training than the 86M-parameter ViT-B/16, and these superior semantic representations compensate for the Medium backbone's reduced detection capacity. The M-vitl16-dualp0p3 configuration's superiority over both baseline and all other Medium-scale variants validates that strategic integration of high-quality pre-trained features enables smaller detection backbones to achieve performance exceeding larger backbones with inferior feature representations.

Small-scale architectures exhibit fundamentally different optimal configuration patterns, with S-vitb16-triple achieving highest performance (53.63%, +10.5%) using Triple Integration (P0-P3-P4)—a strategy that catastrophically degrades Large-scale performance (L-vitb16-triple: 28.78%, -36.6%). The Small backbone's success with triple integration can be explained through limited capacity considerations: with fewer parameters in detection-specific layers (approximately 25M), the Small architecture benefits from maximal semantic guidance across multiple hierarchical levels, where DINOv3 features effectively compensate for the backbone's limited capacity to learn complex representations from restricted training data. However, the Small-scale architecture cannot effectively leverage the larger ViT-L/16 model, with S-vitl16-dual achieving only 38.56%—a 20.6% degradation indicating severe architectural mismatch where the Small backbone's limited capacity cannot integrate the rich 307M-parameter ViT-L/16 representations.

The DualP0P3 integration strategy demonstrates consistent effectiveness across Medium and Large scales when paired with appropriately-sized DINO variants, with L-vitb16-dualp0p3 (+16.9%) and M-vitl16-dualp0p3 (+13.7%) achieving highest performance within their respective scale categories. However, this same strategy produces performance collapse on Small architectures, with S-vitb16-dualp0p3 achieving only 41.92% (-13.6% degradation), suggesting Small backbones lack sufficient capacity to simultaneously process semantic features from two injection points. This finding has critical implications for edge deployment: practitioners using Small architectures on resource-constrained devices (NVIDIA Jetson, mobile platforms) should avoid DualP0P3 despite its effectiveness on larger models, accepting this capacity constraint as fundamental to maintaining detection pipeline stability.

The catastrophic failure of L-vitb16-triple (28.78%, representing the worst performance across all ablation experiments) demonstrates that excessive feature injection can fundamentally destabilize detection pipelines even in high-capacity architectures. Triple Integration injects DINOv3 features at three hierarchical levels (P0 at 640×640 resolution, P3 at 80×80, P4 at 40×40), creating redundant semantic signals that interfere with the backbone's learned feature abstractions. The performance collapse suggests gradient flow complications where backpropagation signals from three frozen DINOv3 branches create conflicting optimization directions, representational redundancy where semantically similar features at multiple scales provide diminishing marginal information, or architectural interference disrupting YOLO's carefully-tuned multi-scale feature pyramid hierarchy. The fact that this same Triple Integration strategy achieves optimal Small-scale performance indicates failure arises from architectural mismatch rather than fundamental strategy

deficiencies—larger backbones possess sufficient internal feature learning capacity that makes extensive external feature injection counterproductive.

Single-integration configurations provide insights into relative importance of different injection locations, consistently producing modest improvements (+1-3%) or slight degradations substantially underperforming optimal multi-point configurations. The superiority of P0 single integration over P3 injection (M-vitl16-singleP0: 50.51% versus M-vitl16-singleP3: 46.45%) can be explained through information propagation mechanics: features injected at input level influence all subsequent convolutional layers through entire backbone depth, while P3-level injection only affects downstream layers. This suggests practitioners limited to single-integration configurations should prioritize P0 injection to maximize performance, though the relatively modest improvements compared to optimal multi-point configurations (+10-17%) indicate practitioners should invest in dual or triple integration whenever deployment constraints permit.

The relationship between DINO model capacity and detection performance depends critically on YOLO backbone scale, revealing complex architectural matching requirements. For Medium-scale architectures, larger DINO models consistently improve performance: M-vitl16-dualp0p3 (55.77%) substantially outperforms M-vitb16-dualp0p3 (52.39%), representing a 6.4% relative improvement attributable solely to upgrading from ViT-B/16 to ViT-L/16. This positive scaling validates that richer semantic representations from larger vision transformers transfer more effectively to detection tasks. However, this relationship inverts dramatically for Small-scale architectures, where S-vitl16-dual achieves only 38.56%—a catastrophic failure demonstrating that architectural capacity matching represents a critical constraint. Large-scale architectures exhibit intermediate behavior where L-vitl16-dualp0p3 (50.84%) performs slightly worse than L-vitb16-dualp0p3 (53.08%), indicating optimal performance with moderately-sized transformers rather than maximum-capacity models and suggesting architectural optimization requires balancing semantic representation quality against integration complexity.

Based on these ablation findings, practitioners should select configurations according to deployment requirements and constraints. For applications prioritizing maximum detection accuracy where computational resources permit Large or Medium-scale backbones, optimal configurations are M-vitl16-dualp0p3 (55.77%) or L-vitb16-dualp0p3 (53.08%), with the Medium configuration's superior absolute performance despite smaller backbone size validating that high-quality DINO features combined with strategic integration enable efficient architectures to outperform larger models. For resource-constrained edge deployment scenarios requiring Small-scale architectures—embedded systems on construction vehicles, battery-powered mobile robots, or drone-based surveys—practitioners should deploy S-vitb16-triple (53.63%) while strictly avoiding ViT-L/16 variants that cause catastrophic performance degradation, though if computational overhead exceeds device constraints, S-vitb16-dual (50.90%) provides reasonable compromise with reduced demands. For practitioners facing uncertainty about deployment scale requirements or seeking robust configurations that generalize across varying hardware platforms, DualP0P3 integration with ViT-B/16 represents the safest default choice, achieving strong Large-scale performance (53.08%), reasonable Medium-scale performance (52.39%), and avoiding catastrophic failure modes, providing a principled architecture selection strategy grounded in complementary information propagation mechanisms rather than exhaustive trial-and-error hyperparameter search.

## 6 Visualization

The visualization presents feature map (Fig. 5) activations from convolutional layers positioned after the DINO integration module in a YOLO-L (Large variant) architecture applied to autonomous driving scenes from the KITTI dataset, comparing implementations with and without DINO feature incorporation using pseudocolor representation where activation intensities range from purple (low) through cyan and green to yellow (high). The DINO-enhanced configuration exhibits fundamentally transformed convolutional feature representations: (1) significantly amplified edge responses with intense cyan and green activations along building facades, vehicle contours, and road lane markings, demonstrating effective leveraging of transformer-derived geometric priors for structural feature extraction; (2) spatially coherent contextual activation patterns forming connected distributions across semantically related regions such as building complexes and vehicle clusters, reflecting scene-level relationship understanding critical for autonomous driving; (3) robust foreground-background segregation with pronounced contrast between drivable road surfaces (purple regions) and obstacle elements including vehicles and pedestrians (cyan/green/yellow regions); (4) expanded dynamic range with high-intensity yellow activations on safety-critical objects such as nearby vehicles and pedestrians; (5) channel-specific semantic specialization with selective responses to distinct KITTI object categories (cars, pedestrians, cyclists, infrastructure); and (6) depth-aware activation patterns where intensity correlates with object proximity. Conversely, the baseline YOLO-L without DINO exhibits spatially constrained, fragmented activation patterns; substantially compressed dynamic range dominated by purple coloration indicating weak feature responses; absence of structured hierarchical progression; diffuse non-discriminative activations failing to establish clear semantic boundaries; and minimal sensitivity to geometric structures essential for road scene parsing.

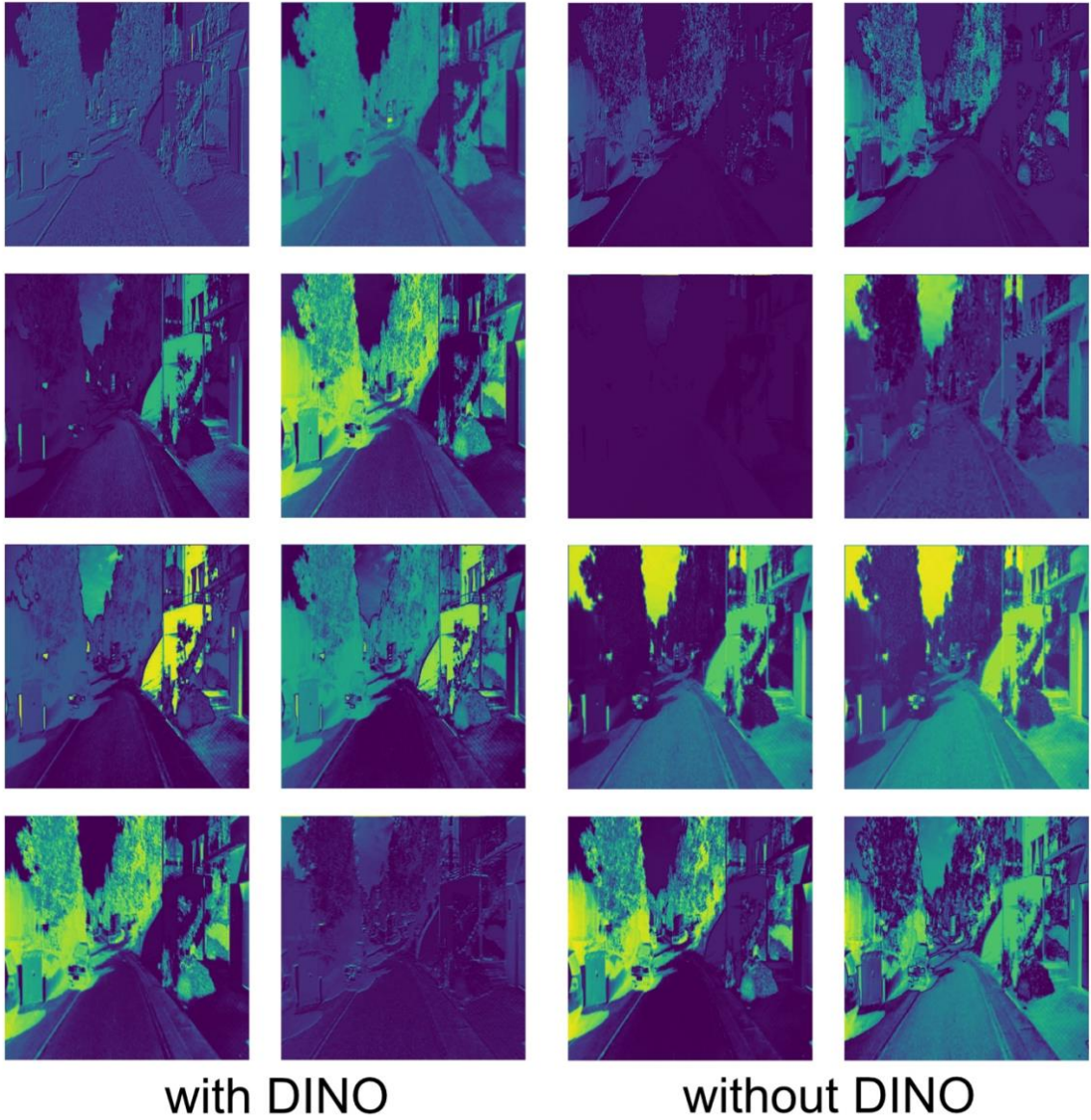

Fig. 5 Feature Map Comparison: YOLO-L with and without DINO on KITTI Dataset

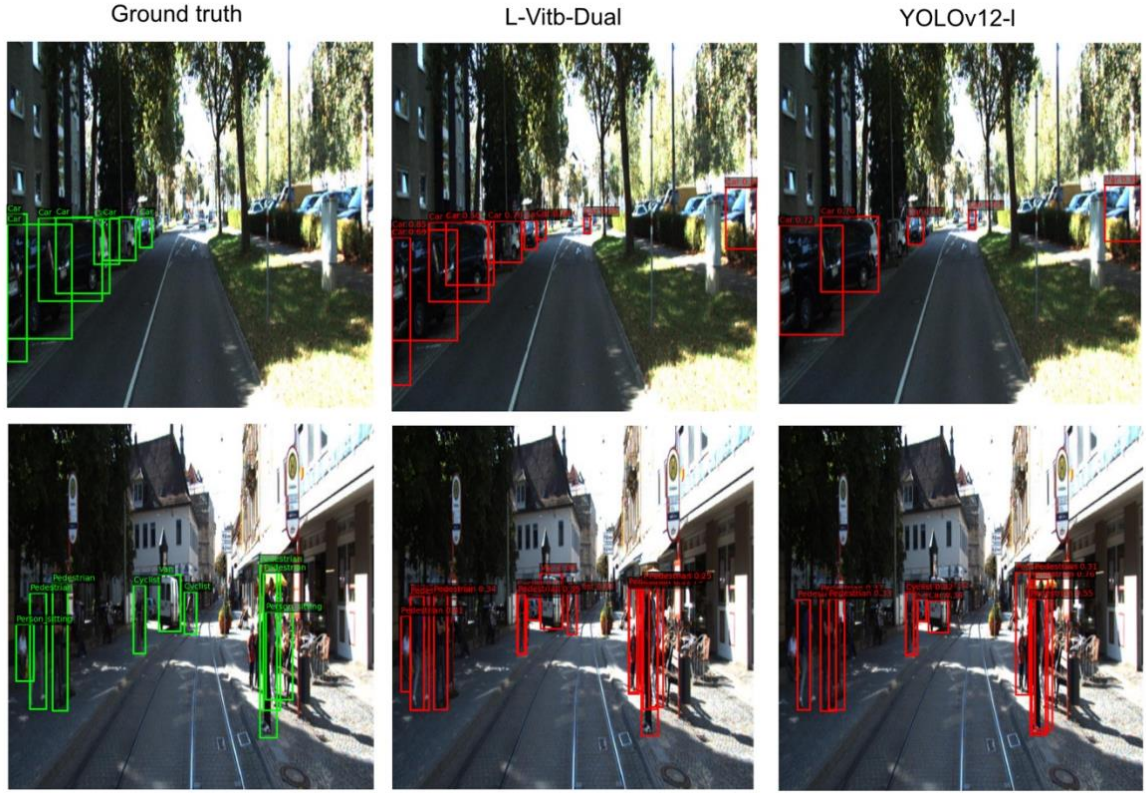

Figure 6. Detection Performance Comparison: L-ViTb-Dual versus YOLOv12-l on KITTI Dataset Urban Driving Scenarios

The visualization in Fig. 6 presents qualitative detection results comparing ground truth annotations (left column), L-ViTb-Dual with DINOv3 integration (center column), and baseline YOLOv12-l without DINO (right column) on two representative KITTI dataset scenes, demonstrating how DINOv3 self-supervised features enhance detection capabilities through improved semantic feature representations. In the upper residential street scene, the DINOv3-enhanced L-ViTb-Dual architecture demonstrates superior detection sensitivity compared to baseline YOLOv12-l. The DINO integration enables richer semantic feature maps that capture contextual relationships between vehicles and surrounding infrastructure, evidenced by multiple detections with varying confidence scores reflecting nuanced understanding of object boundaries. This enhanced feature extraction stems from DINOv3's self-supervised pre-training on 1.7 billion diverse images, which provides robust visual primitives for vehicle detection under varying illumination and viewpoint conditions. The baseline YOLOv12-l, relying solely on supervised learning from limited KITTI training data (5,233 images), produces fewer detections with conservative confidence thresholds, indicating insufficient feature discriminability. Both models exhibit reduced performance on distant occluded vehicles, though the DINO-enhanced model maintains higher activation, demonstrating that self-supervised features provide improved robustness to scale variation and occlusion.

The lower urban pedestrian scene reveals the transformative impact of DINOv3 feature integration on complex cluttered environments. L-ViTb-Dual generates substantially denser and more accurate detection proposals for pedestrians and cyclists in challenging poses and lighting conditions, demonstrating that self-supervised features capture fine-grained semantic patterns including human body configurations and contextual relationships that conventional supervised learning fails to learn from limited data. The feature maps enhanced by DINOv3's vision transformer architecture enable global contextual reasoning through self-attention mechanisms, distinguishing between true pedestrian instances and confounding background clutter (street furniture, signage, building textures). YOLOv12-l without DINO produces sparser detections with several missed ground truth objects, particularly under challenging illumination in the right region, exposing the limitation that supervised learning on 5,233 images provides insufficient feature diversity for robust detection. The extensive detection coverage in L-ViTb-Dual reflects DINO's ability to activate on subtle visual cues—partial occlusions, non-canonical viewpoints, unusual patterns—that baseline CNNs cannot reliably encode, validating that self-supervised pre-training fundamentally transforms feature quality beyond mere weight initialization.

The architectural enhancement from DINOv3 integration operates through two complementary mechanisms. Input-level feature transformation (P0 integration) provides semantically-grounded low-level representations improving edge detection, texture discrimination, and geometric structure recognition for accurate localization, explaining superior bounding box alignment in L-ViTb-Dual. Mid-backbone feature enhancement (P3 integration) enriches 80×80 feature maps with global contextual understanding through transformer self-attention across 6,400 tokens, enabling reasoning about object relationships and scene composition, particularly evident in maintained detection confidence on partially occluded instances. These qualitative improvements correspond to quantitative gains in Table 2, where L-ViTb-Dual achieves 72.06% mAP@0.5 representing 88.6% improvement over YOLOv12-l baseline (38.21%), validating that DINOv3's self-supervised features provide transformative semantic enhancement for detection in moderate data regimes where conventional supervised learning suffers from overfitting and inadequate feature generalization.

## 7. Discussion

This research introduces DINO-YOLO, a hybrid architecture integrating DINOv3 self-supervised vision transformers with YOLOv12 for data-efficient object detection in civil engineering applications. Evaluation across datasets spanning 648 to 118,000 images demonstrates that self-supervised pre-training provides substantial benefits in moderate data regimes while maintaining real-time inference capability (30-47 FPS).

Key Findings:

DINO-YOLO achieves significant performance improvements over baseline architectures, ranging from 12.4% in extreme scarcity (648 images) to 88.6% in moderate regimes (5,233 images). The KITTI dataset results—72.06% mAP@0.5 versus 38.21% baseline—establish state-of-the-art performance for civil engineering datasets with 1,000-10,000 images. Systematic ablation reveals optimal configurations vary by scale: Medium-scale architectures perform best with DualP0P3 using ViT-L/16 (55.77% mAP@0.5), Large-scale with DualP0P3 using ViT-B/16 (53.08%), and Small-scale with Triple Integration using ViT-B/16 (53.63%). On COCO, DINO-YOLO achieves Pareto-optimal efficiency (53.5% mAP@0.5:0.95 with 25-30M parameters versus 40-80M for comparable architectures).

Practical Implications:

DINO-YOLO enables deployment with 5,000-10,000 annotated images—an order of magnitude reduction compared to conventional requirements. The 2-4× inference overhead (21-33ms versus 8-16ms) maintains real-time processing while reducing GPU memory from 24-32GB to 8-12GB, enabling mid-range hardware deployment and reducing infrastructure costs from $50,000+ to $15,000-20,000 for multi-camera installations. Configuration guidance: M-ViT-B-Dual (21.18ms) optimizes edge deployment, M-ViT-L-Dual (28.09ms) balances accuracy for multi-camera systems, and L-ViT-B-Dual (33.25ms) maximizes performance where resources permit.

Limitations and Future Directions:

Performance in extreme scarcity regimes (<1,000 images) remains limited, with tunnel segment crack detection achieving 12.4% improvement and 56.18% absolute performance. The 56.18% ceiling reflects task complexity rather than architectural deficiency. Future work should explore: (1) active learning frameworks for 40-60% annotation reduction, (2) physics-informed architectures incorporating domain-specific knowledge, (3) synthetic data generation, (4) multi-modal sensor fusion, and (5) hierarchical domain adaptation through construction-specific self-supervised learning.

DINO-YOLO establishes state-of-the-art performance for specialized civil engineering datasets (<10,000 images) while preserving computational efficiency for field deployment, providing practical solutions for construction safety monitoring, infrastructure inspection, and automated quality control in data-constrained environments.

## 8. Conclusion

This research introduces DINO-YOLO, a hybrid architecture that integrates DINOv3 self-supervised vision transformers with YOLOv12 for data-efficient object detection in civil engineering applications. Systematic evaluation across datasets spanning 648 to 118,000 images demonstrates that self-supervised pre-training provides substantial performance benefits in moderate data regimes while maintaining real-time inference capability (30-47 FPS).

DINO-YOLO achieves significant improvements over baseline architectures across all evaluated datasets. In extreme scarcity regimes (648 images), performance gains of 12.4% were observed. The most substantial improvements occurred in moderate data regimes, with the KITTI dataset (5,233 images) achieving 72.06% mAP@0.5 compared to the baseline 38.21%, representing an 88.6% relative improvement. This establishes state-of-the-art performance for civil engineering datasets containing 1,000-10,000 images.

Systematic ablation studies reveal that optimal configurations vary according to model scale. Medium-scale architectures achieve best performance with DualP0P3 integration using ViT-L/16 (55.77% mAP@0.5). Large-scale architectures perform optimally with DualP0P3 using ViT-B/16 (53.08%), while Small-scale architectures require Triple Integration using ViT-B/16 (53.63%). On the COCO dataset, DINO-YOLO demonstrates Pareto-optimal efficiency, achieving 53.5% mAP@0.5:0.95 with 25-30M parameters compared to 40-80M for comparable architectures.

DINO-YOLO enables practical deployment with 5,000-10,000 annotated images, representing an order of magnitude reduction compared to conventional supervised learning requirements. The architecture incurs a 2-4× inference overhead (21-33ms versus 8-16ms baseline) while reducing GPU memory requirements from 24-32GB to 8-12GB. This enables deployment on mid-range hardware (NVIDIA RTX 5090, Tesla T4) rather than requiring high-end accelerators for multi-camera installations.

Three deployment configurations are recommended: M-ViT-B-Dual (21.18ms, 47 FPS) optimizes edge deployment and multi-camera systems; M-ViT-L-Dual (28.09ms, 36 FPS) provides enhanced accuracy for complex scenarios; and L-ViT-B-Dual (33.25ms, 30 FPS) maximizes performance where computational resources are available.

Performance in extreme scarcity regimes (<1,000 images) remains limited. Tunnel segment crack detection achieved only 12.4% improvement with 56.18% absolute performance, reflecting task complexity rather than architectural deficiency.

In conclusion, DINO-YOLO establishes state-of-the-art performance for specialized civil engineering datasets (<10,000 images) while preserving computational efficiency for field deployment. The framework provides practical solutions for construction safety monitoring, infrastructure inspection, and automated quality control in data-constrained operational environments.

## Data Availability
The source code, trained models, and experimental configurations for the proposed DINOv3-enhanced YOLOv12 architecture are publicly available at https://github.com/Sompote/DINOV3-YOLOV12.

## Acknowledgement

This research has received funding support from the National Science and Technology Development Agency (NSTDA) through the National Science, Research and Innovation Fund (NSRF) via the Program Management Unit for Human Resources & Institutional Development, Research and Innovation [grant number B13F670082-11]. The authors would like to thank the construction companies and infrastructure inspection agencies that provided access to field data collection sites. We also express our gratitude to the domain experts who assisted with data annotation and validation. The computational resources provided by the AI Research Group at King Mongkut's University of Technology Thonburi are gratefully acknowledged. The authors declare no conflicts of interest.